\documentclass[letterpaper, 10 pt, conference]{ieeeconf}  
\usepackage{hyperref}
\usepackage{caption}
\usepackage{subcaption}
\usepackage{xcolor}
\usepackage{tikz}
\usetikzlibrary{positioning}
\definecolor{myred}{rgb}{0.9, 0.1, 0.1}

\IEEEoverridecommandlockouts                      
\title{\LARGE \bf
A 3D Mixed Reality Interface for Human-Robot Teaming
}

\author{Jiaqi Chen*$^{1}$, Boyang Sun*$^{1}$, Marc Pollefeys$^{1,2}$, Hermann Blum$^{1}$\\
\thanks{*Equal contribution}
\thanks{$^{1}$ETH Zürich, Switzerland}
\thanks{$^{2}$Microsoft Mixed Reality Lab, Zürich, Switzerland}
\vspace{-5mm}
}

\usepackage{amsmath}
\DeclareMathOperator*{\Avg}{AVG}
\begin{document}

\twocolumn[{%
\renewcommand\twocolumn[1][]{#1}%
\maketitle
\thispagestyle{empty}
\vspace{-0.4cm}
\resizebox{\linewidth}{!}{%
\begin{tikzpicture}[style={outer sep=0,inner sep=0}]
\node[label={drag\strut...}] (grab1) {\includegraphics[width=.3\textwidth]{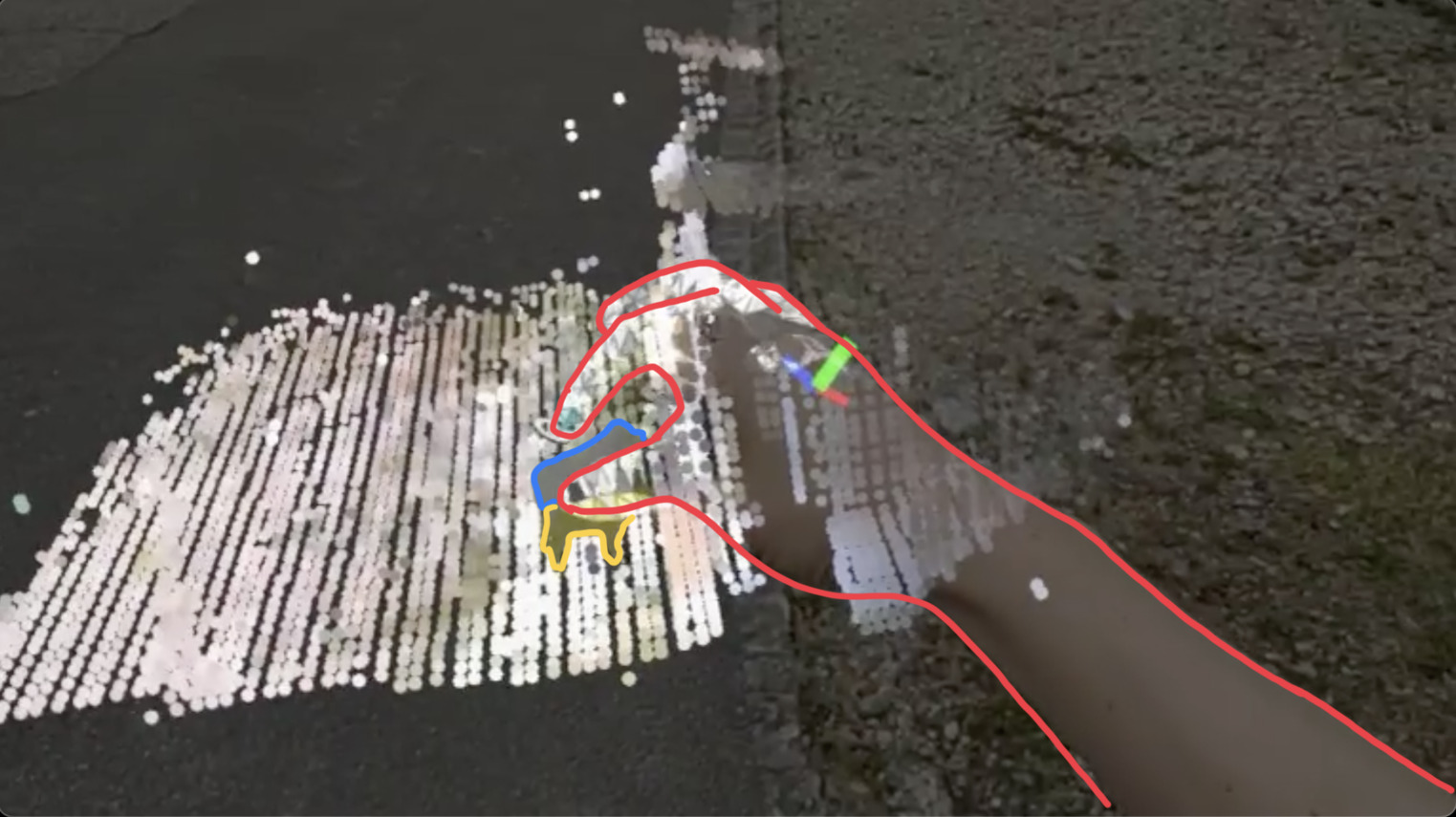}};
\node[label={...and \strut drop...},right=1mm of grab1] (grab2) {\includegraphics[width=.3\textwidth]{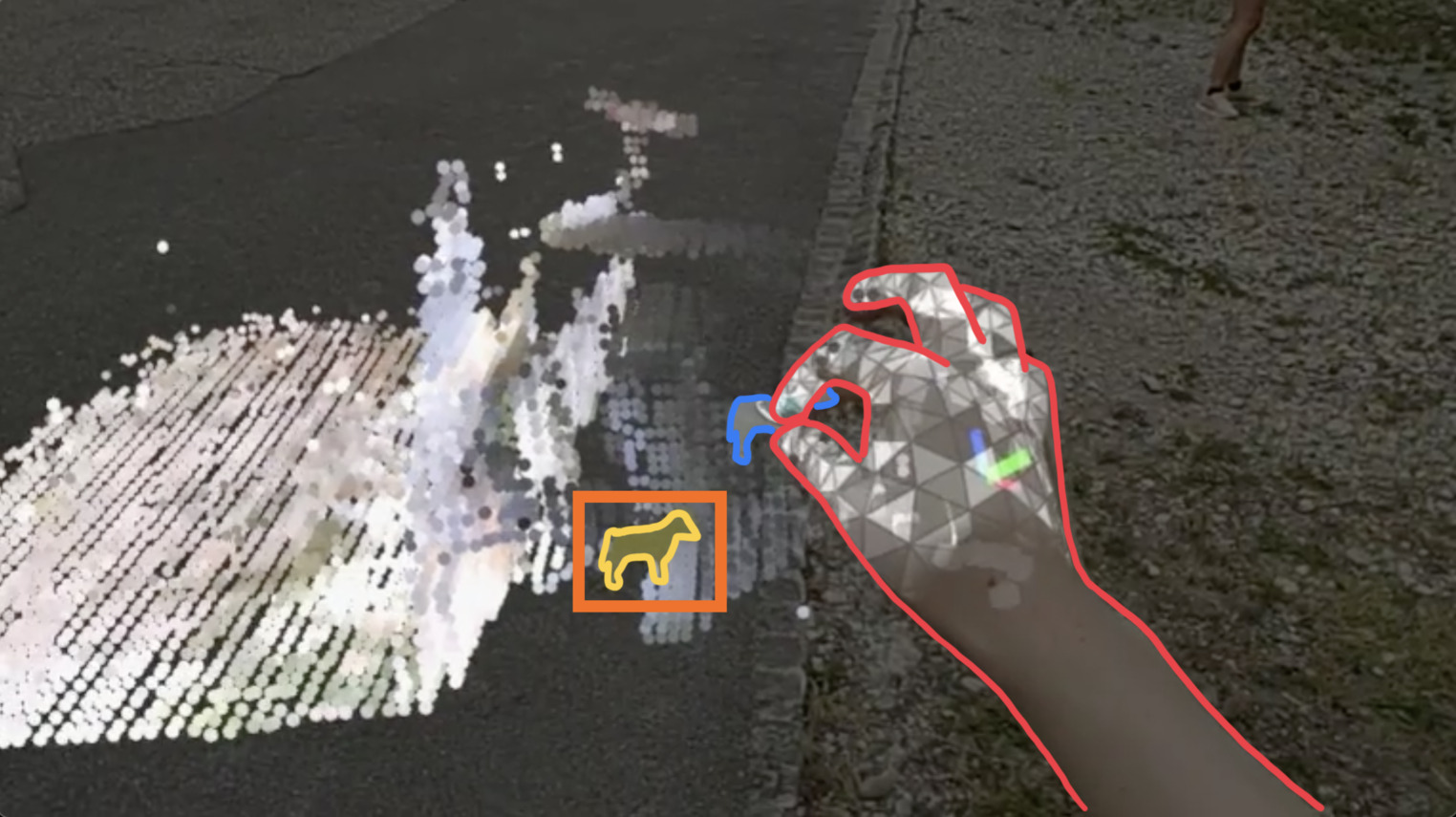}};
\node[label={...to \strut move the robot.},right=1mm of grab2] (grab3) {\includegraphics[width=.3\textwidth]{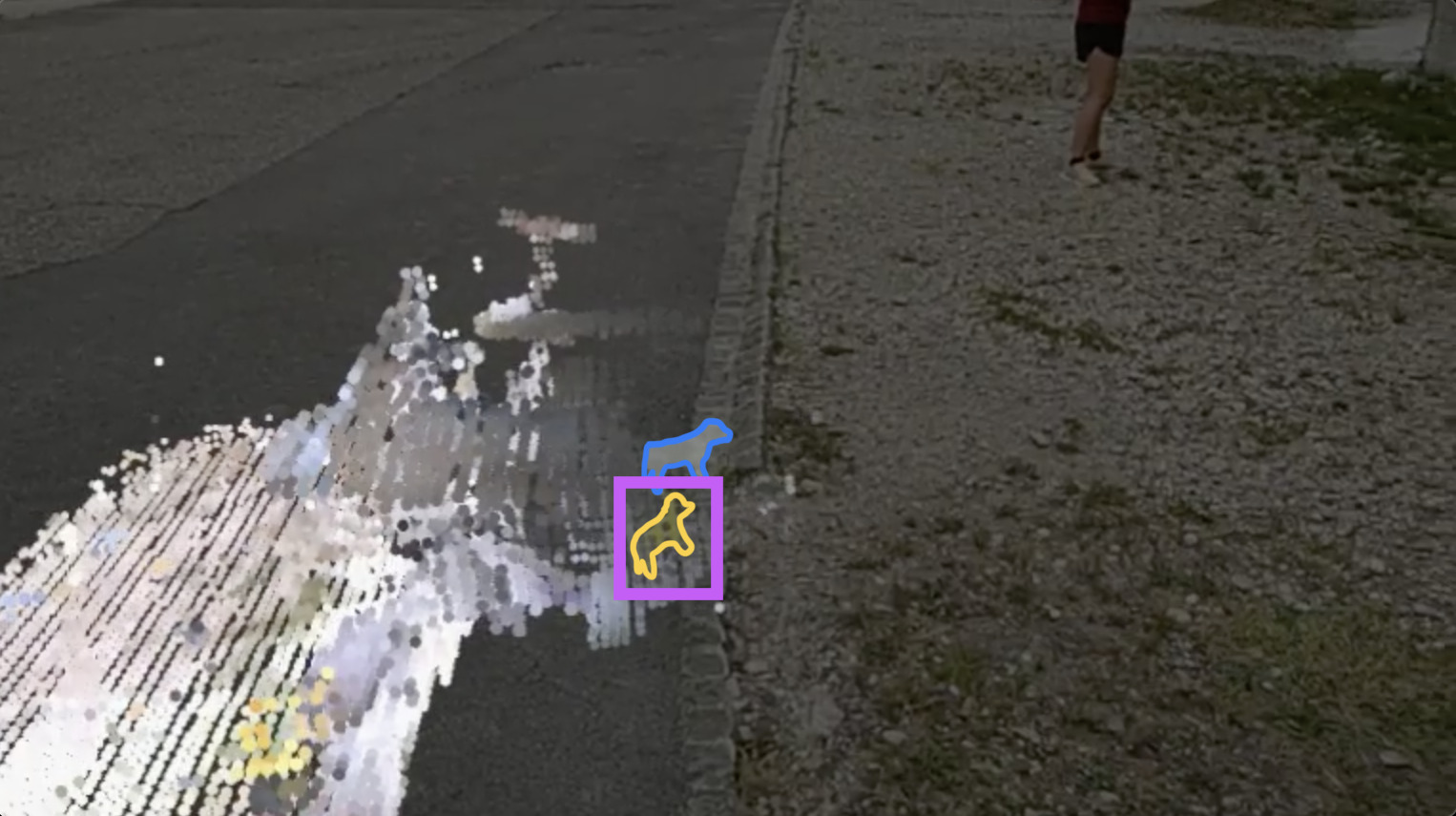}};
\node[below=1mm of grab1] (view1) {\includegraphics[width=.3\textwidth]{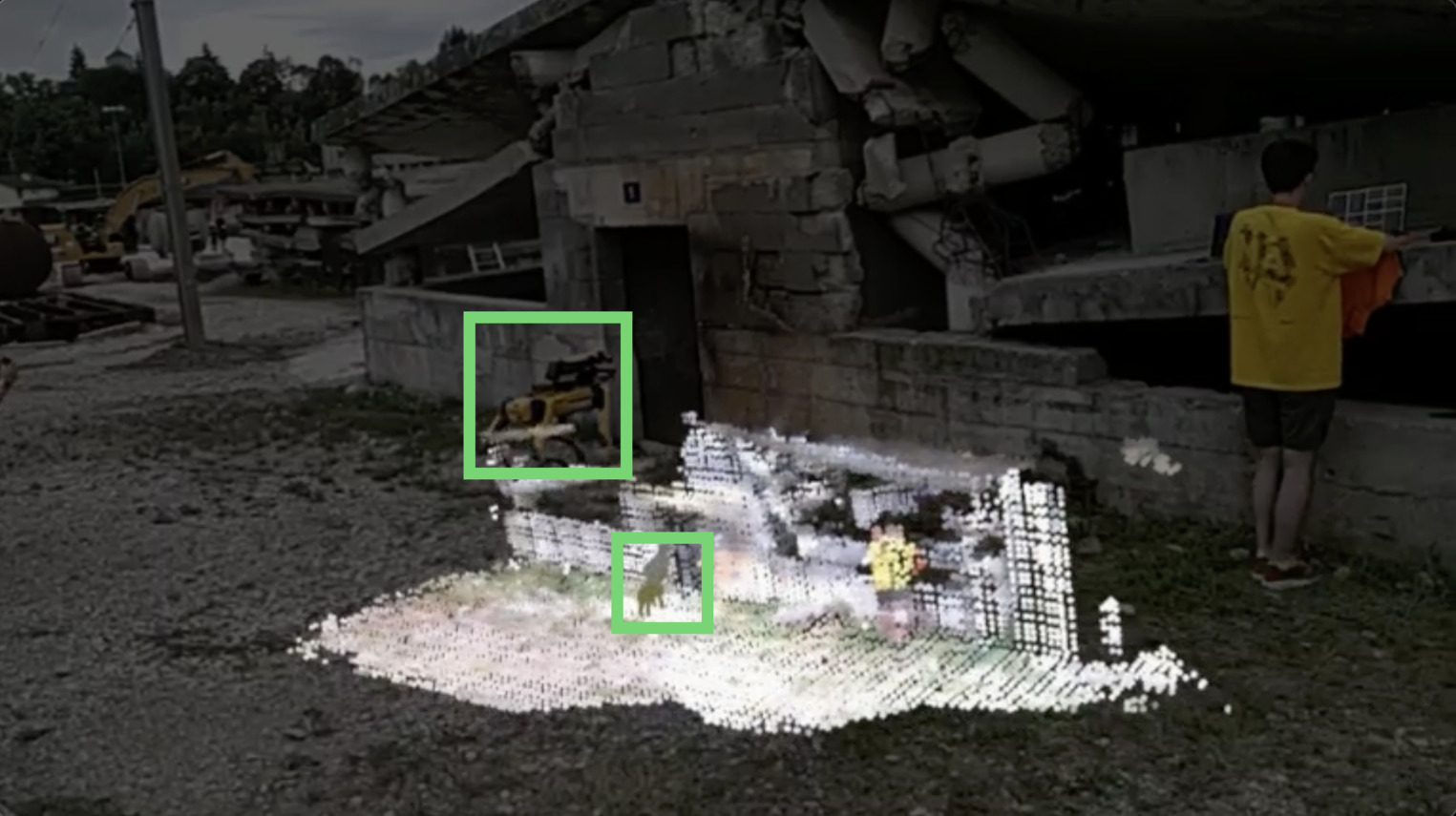}};
\node[right=1mm of view1] (view2) {\includegraphics[width=.3\textwidth]{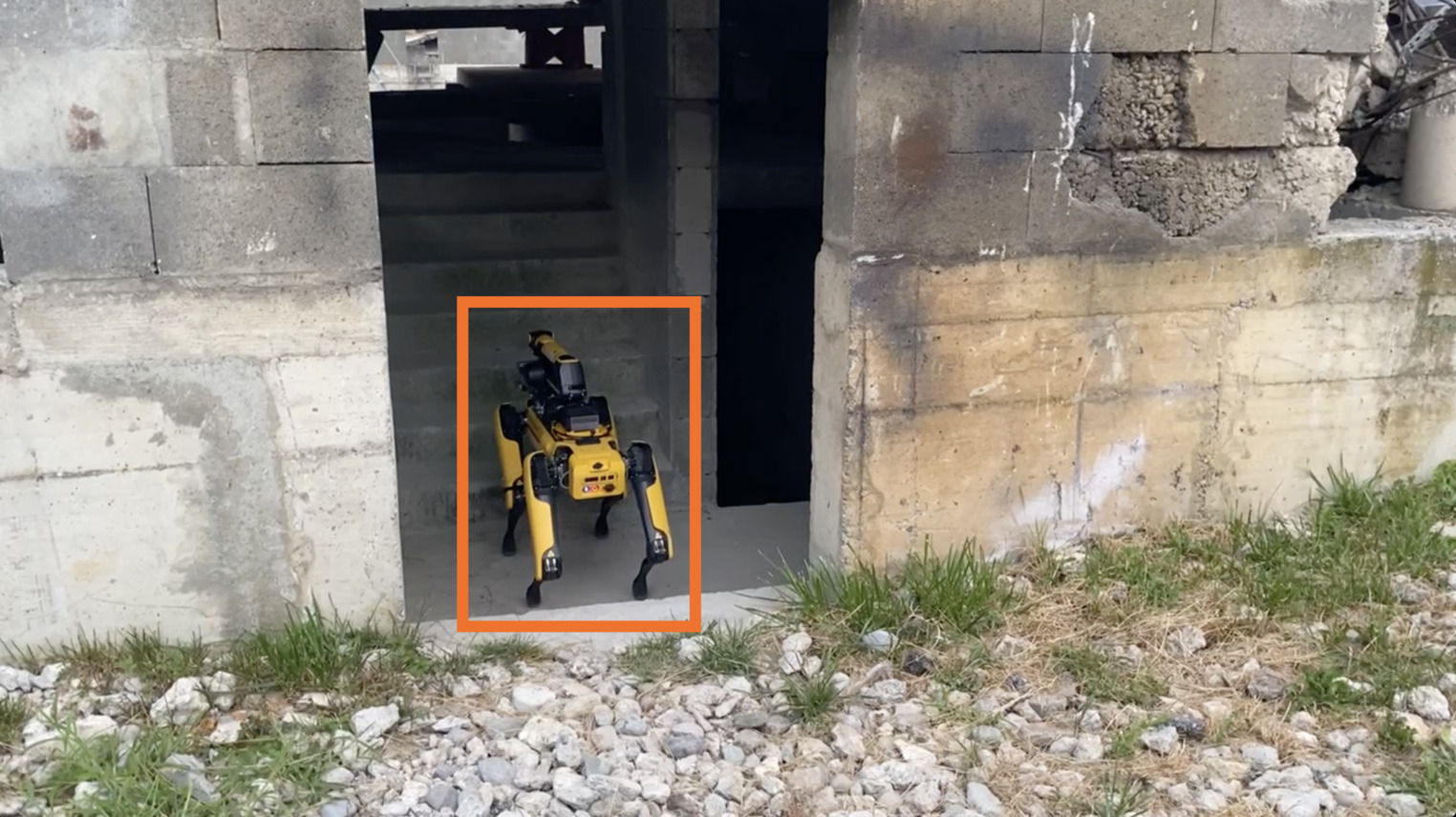}};
\node[right=1mm of view2] (view3) {\includegraphics[width=.3\textwidth]{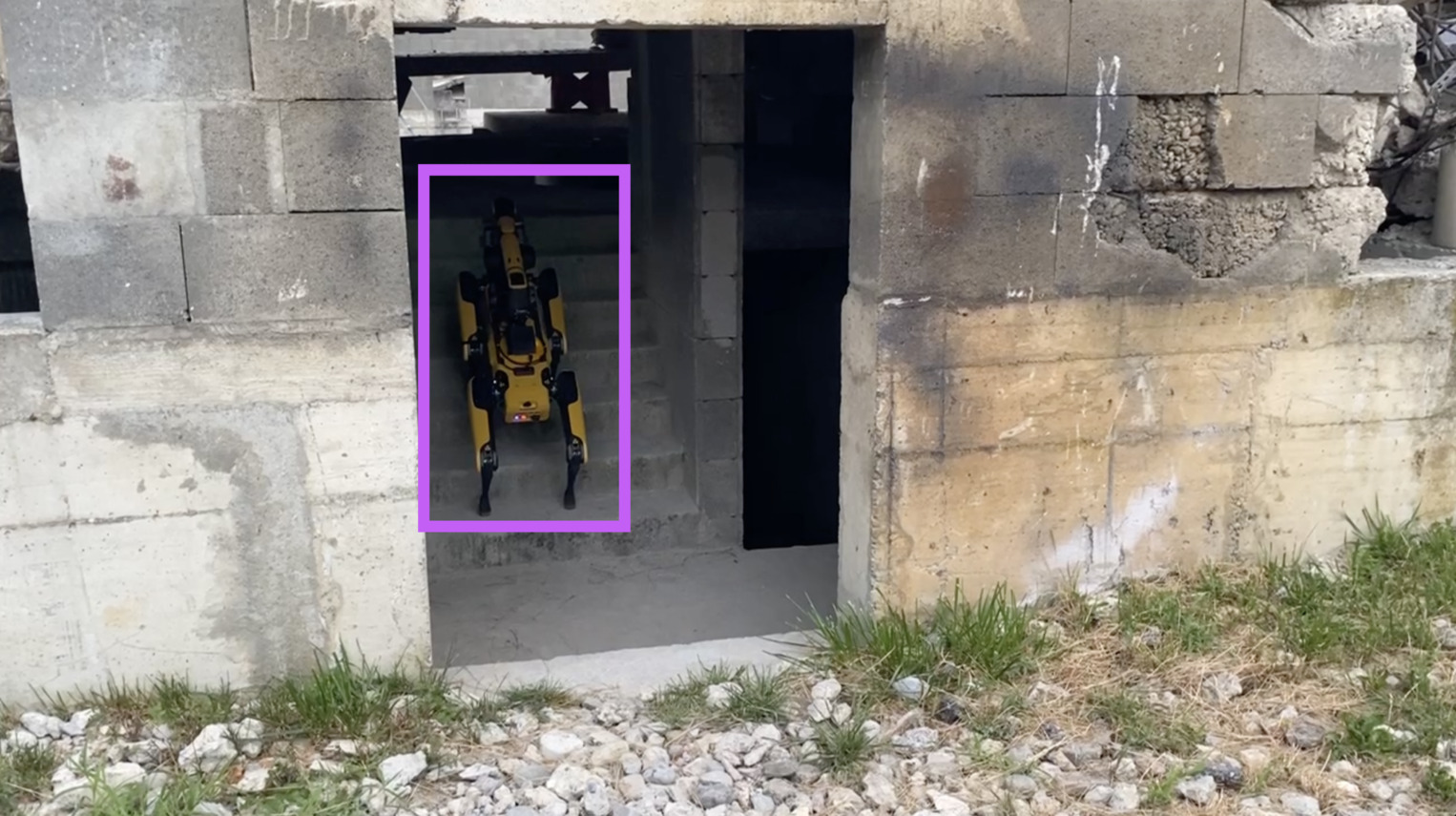}};

\node[left=1mm of grab1]{\rotatebox{90}{Operator View\strut}};
\node[left=1mm of view1]{\rotatebox{90}{Environment View\strut}};
\end{tikzpicture}%
}
\vspace{-15px}
\captionof{figure}{\textbf{
Demonstration of Our System}
The top row shows a first-person view from the HoloLens of controlling a robot through drag-and-drop. The bottom row shows how the robot moves accordingly---from the operator's viewpoint in the first frame, then from an external camera as the robot moves into the building.
Video at \href{https://youtu.be/H3IA5FXnFX8}{\texttt{youtu.be/H3IA5FXnFX8}}.
}
\label{fig:arche_controller_mode}
}]
\thispagestyle{empty}
\pagestyle{empty}

\begin{abstract}

This paper presents a mixed-reality human-robot teaming system. 
It allows human operators to see in real-time where robots are located, even if they are not in line of sight. The operator can also visualize the map that the robots create of their environment and can easily send robots to new goal positions.
The system mainly consists of a mapping and a control module. The mapping module is a real-time multi-agent visual SLAM system that co-localizes all robots and mixed-reality devices to a common reference frame.
Visualizations in the mixed-reality device then allow operators to see a virtual life-sized representation of the cumulative 3D map overlaid onto the real environment. 
As such, the operator can effectively ``see through" walls into other rooms. 
To control robots and send them to new locations, we propose a drag-and-drop interface. An operator can grab any robot hologram in a 3D mini map and drag it to a new desired goal pose. We validate the proposed system through a user study and real-world deployments. We make the mixed-reality application publicly available at \href{https://github.com/cvg/HoloLens_ros}{\texttt{github.com/cvg/HoloLens\_ros}}.

\end{abstract}
\let\thefootnote\relax\footnotetext{\hspace{-\parindent}* equal contribution\\
$^{1}$ Computer Vision and Geometry Lab, ETH Zürich\\
$^{2}$ Microsoft Mixed Reality \& AI Lab, Zürich\\
This work was partially supported by the ETH Foundation Mobility Initiative and the SUTD Beta \& Gamma Draconis Programs.
}

\section{INTRODUCTION}


Robots have successfully conquered many 2D worlds as they trim our greens, vacuum our apartments, or guide us through museums.
More recently, the improved physical capabilities of robotic systems enabled the deployment of drones, legged robots, and cable-driven robots, amongst others, in a wide range of 3D physical scenarios.
However, the challenge lies in effectively maneuvering these robots in a 3D space. The most common human-robot interfaces by far are 2D screens that, however, can only provide a limited perspective into three-dimensional environments, and accept control input solely as 2D pixel coordinates. Traditional solutions that try to bridge the gap into direct 3D controls often involve splitting input and output, such as using specialized 3D pointers, multiple joysticks, or other form of physical controls \cite{teleop1_robot_teleop_ar_virtual_surrogates,teleop2_sauer2009_towards_a_predictive_mr_ui_for_mobile_robot_teleop,teleop3_bejczy2020_mr_interface_for_improving_mob_manip,teleop4_sauer2010_mr_ui_for_mobile_robot_teleop_in_adhoc,teleop5_rob_teleop_based_on_mr,teleop6_legged_manip}.


Mixed-reality (MR) headsets have surged in capability, prompting initial explorations into their potential for robot operation~\cite{walker2022virtual}. This research delves into the untapped potential of harnessing the 3D input and output capabilities inherent in virtual and mixed reality for human-robot teaming. Due to the nature of mixed reality, operators can also see holograms overlaid onto the real world~\cite{reardon2019communicating}. This feature is beneficial for scenarios such as search and rescue, where operators need visuals on what their robots are doing behind an occluded space. Such benefits of MR interfaces have been investigated in some existing works, but still have limits in either the visualization or operation side.
This work demonstrates a 3D, real-world scale, mixed-reality visualization system which spatially registers all operators and robots without any external alignment and lets operators ``see through walls." This is enabled by a real-time multi-agent SLAM system, combined with dense reconstruction, gravity priors and reference frame averaging. Furthermore, we propose a drag-and-drop interface to intuitively operate \textit{multiple} robots in \textit{3D}. A user study shows that when operating multiple robots in a multi-floor environment, MR is a significantly faster interface than operating 3D capable robots on a 2D screen. Finally, we perform real-world validations of our system. Through this investigation, new frontiers emerge for enhancing teleoperation and teaming with robots in complex 3D spaces. In summary, our contributions are
\begin{itemize}
    \item A novel drag-and-drop interface to intuitively operate \textit{multiple} robots in \textit{3D}.
    \item A 3D mapping and visualization system that co-localizes robots and operators and enables the latter to ``see through walls."
    \item A user study for the effectiveness of the drag-and-drop-based robot control interface.
    \item Real-world validations of the described system.
\end{itemize}

\section{Related Works}


The fusion of mixed reality and robotics has emerged as a popular and exciting research area. Most works in this field focus on intuitive robot controls to enable simultaneous perception and interaction with the environment. Concise reviews are provided in \cite{Suzuki_2022} and \cite{walker2022virtual}. We highlight several recent and most relevant papers.  

\subsection{Operation through Virtual/Mixed Reality}
Researchers have been noticing the advantages of using 3D interfaces to overcome the limitations of 2D displays. This concept has been applied in commanding robotic arms \cite{9561105,naceri2021vicarios}, in which virtual reality interfaces are proposed for controlling robot arms to manipulate objects. \cite{9561105} suggests an interface that directly links the desired payload position from the operator to the joint commands of multiple robot arms. Similar ideas have also been applied to mobile robots. In \cite{wu2018omnidirectional}, the authors propose an MR interface to detect hand gestures and map them to commands for an omnidirectional ground robot. This however limits the operator to only perform first-person operations. In \cite{8968598}, a teleoperation system is proposed, which enables the operator to send commands to a mobile robot to do live exploration. The system also builds a 3D reconstruction of the explored area, so operators are no longer constrained to a first-person viewpoint. However, their system only works with a single robot. And since it uses Virtual Reality, the operator loses almost all visual information from the real world. \cite{cruz2023mixed} builds an MR interface for commanding ground robots. Operators command robots by setting holographic waypoints in the real world, but they cannot set waypoints beyond their line of sight. Similarly, in \cite{9681715}, the authors design a system to perform several collaborative tasks, such as mission planning for inspection, gesture-based control, and teleoperation. The operator is able to set waypoints in 3D space for the robot to follow. However, demonstrations in the paper are in a space with flat ground, only validating 2D tasks, and not tasks that involve operation in multi-floor environments. In addition, the operator must set the waypoints manually, hindering fast operation or command beyond line of sight. As one of the first such interfaces for multiple-robot collaboration, \cite{10161412} proposes an MR interface to command a ground robot swarm, but only on a 2D plane. To conclude, existing works in this field have the limitations that they, a) only support first-person viewpoint operation, b) only support 2D tasks, c) are not able to operate beyond line of sight, or d) do not capture real-world information in real-time.

\subsection{Human-Robot Teaming for Spatial Understanding}
Besides a large interest in using virtual reality to provide a more intuitive interface to remote operators, there is another group of previous work investigating mixed/augmented reality to provide \textit{spatial context} to humans collaborating with robots in a shared space. Streaming the robot camera is a straightforward way \cite{wu2018omnidirectional,hedayati2018improving,cruz2023mixed}. However, it provides very little spatial information to the operator. The authors of \cite{erat2018drone} explore the idea of using a globally localized robot map and an MR headset to let robot operators ``see through walls" into the environment. But their system fully relies on an external optical tracking system. \cite{reardon2018come,reardon2019communicating,sutd_video} go a step further by aligning the MR headset and the robot into a single base frame. Thus, they are able to visualize the robot poses and trajectories, as well as a 2D map, in an MR device in the field. These works demonstrate that such co-localization and visualization modules give a team of a human and robots improved spatial awareness to perform collaborative tasks, such as exploration or search and rescue. However, their alignment methods require good initial manual alignment or an external tracking device. The representations that are used for visualization in these works are 2D occupancy maps, which do not provide a lot of spatial information. Another group of recent works has also focused on further utilizing the perception ability of the robot to help with human operators' situational awareness \cite{reardon2020enabling,mott2021you}. These works focus on environmental perception on the robotics side and send perceived information to operators. In conclusion, research in this direction is limited to, a) Only streaming egocentric information, b) Requiring external co-localization input, c) Visualizing simple 2D occupancy maps.

\section{Methodology}


This section introduces our teaming system and its two major components. The first is a real-time multi-agent visual SLAM pipeline, which we improve with gravity alignment and reference frame averaging procedures. With this SLAM pipeline, we register and track all agents in a global coordinate frame. This global frame enables the second ``mixed-reality" component: visualizing the cumulative map in MR, correctly aligned to the real world. As such, operators can see into adjacent spaces that would otherwise be occluded by walls or obstacles. With all agents localized in a global reference frame, we then add a simple traversability estimation planner to build a drag-and-drop interface to control multiple agents in 6DoF within the HoloLens. 

We highlight that our system works with $N$ number of robots and $M$ number of HoloLenses, but we are resource constrained to one physical robot and one HoloLens for real world tests, and compute constrained to three robots in simulation.

\begin{figure}[t]
    \centering\includegraphics[width=\linewidth]{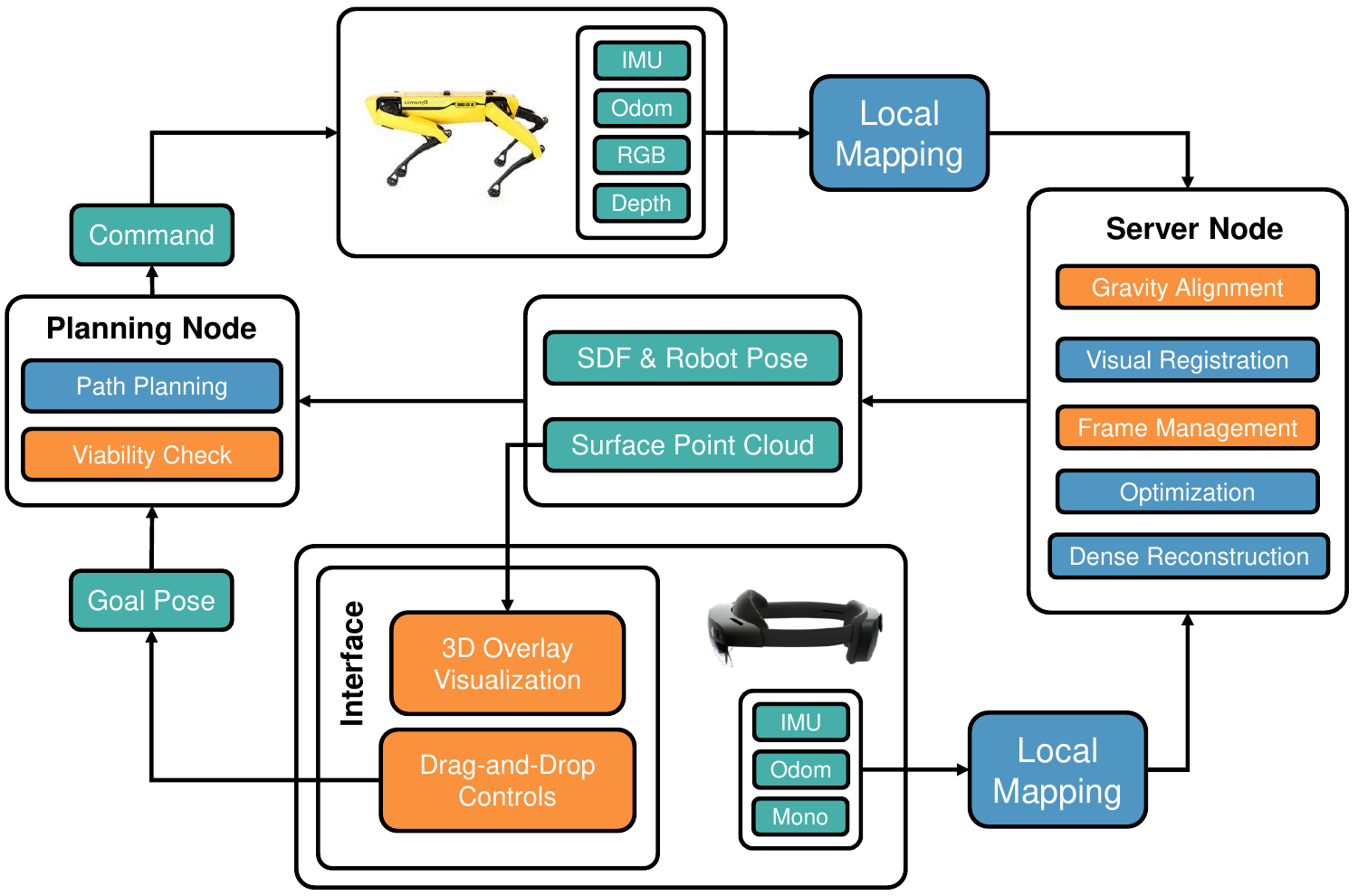}%
    \caption{\textbf{System Overview} The MR device and robot send their respective sensor readings to a local mapping node. The node builds local maps and sends them to the server node. The server builds the global map and sends the 3D point cloud back to the MR device for visualization. It also integrates an SDF (signed distance field) of the environment and communicates with a motion planning node which takes inputs from the operator and commands the robot. Blue components indicate where we integrate existing works, orange components are newly added contributions.} 
    \label{fig:system}
    \vspace{-6mm}
\end{figure}

\subsection{Mapping and Co-localization System}
\label{subsec:mapping-localization}
In this subsection, we introduce our framework for building the map with the robot and MR device, and localizing each agent in the built map.

\subsubsection{Mapping} Our mapping approach is based on \textit{maplab 2.0}~\cite{Cramariuc-2023}. For each agent, we gather camera images, IMU, and self-odometry signals as input. Optionally, a depth frame can be incorporated to construct a dense map. Each agent within the system constructs its own \textit{local map}. This sparse metric map contains keyframes from the agents' sensor, along with the corresponding raw observations of each keyframe and the triangulated 3D landmarks. The base frame of the agent's self-odometry serves as the reference origin for the local map. Observations and pose graphs are divided into submaps by grouping them by time of observation. 

\subsubsection{Co-localization} The co-localization approach involves the alignment and merging of all local maps from each agent into a \textit{unified global map}. The whole process is handled by a centralized \textit{server node}, to which all local maps are transferred as submaps. The first incoming map serves as the base map for alignment, i.e. all future incoming maps will be registered and transformed into the base map. When a local map comes in, we align it to a common convention of the gravity direction, which points in the negative $z$ direction in our convention. Therefore, we average the IMU acceleration vector of all nodes in the pose graph of the incoming map, and rotate the entire map such that the average acceleration is aligned with $z$. We then apply the existing maplab visual registration, which focuses on maximizing the geometric consistency of identical 3D landmarks across different local maps. Landmarks are matched based on nearest neighbor lookups of the descriptors from the incoming map into the base map. This returns a rigid transformation from the incoming map into the common base frame.

Once aligned, we regularly match and merge landmarks observed from multiple agents and run bundle-adjustment over the whole map to further optimize the poses of all agents with respect to the common reference frame.


\begin{figure}
    \centering
    \includegraphics[width=.77\linewidth]{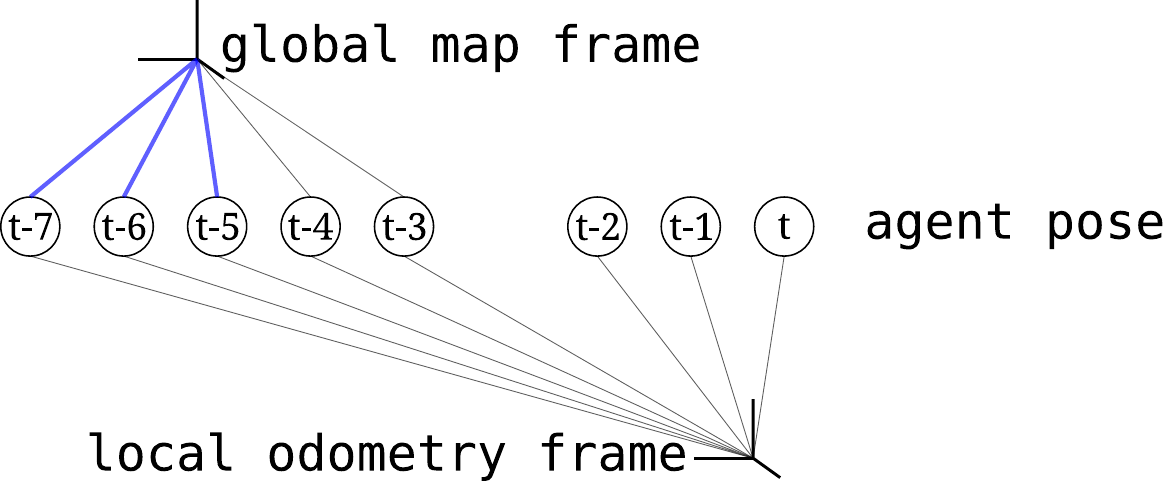}
    \caption{Until the local map is transferred to the global map server and poses are optimized in the map (blue), delay accumulates. To get the most accurate estimate of an agent's current position with respect to the global map frame, we keep a running estimate of $T_{\textrm{map}\leftarrow \textrm{odom}}$ and transform real-time odometry based on this estimate.}
    \label{fig:frame-management}\vspace{-5mm}
\end{figure}

\subsubsection{Reference Frame Management} Once an agent's local map is aligned with the base map, its sensor frames and metric map are expressed within the global map. Consequently, we can query the agent's global location up to the timestamp of the latest incoming submap (in Figure~\ref{fig:frame-management}, that is $t-3$). If we want to e.g. visualize a robot's current position in HoloLens, we, however, need both the robot's current position in the global map frame and a good estimate for where the odometry origin of the HoloLens (which is also the origin of the visualization frame) is with respect to the global map frame. This presents a dilemma: For the most recent poses, we only know $T_{\textrm{odom} \leftarrow \textrm{agent}}$, since they have not yet been integrated into the common map. On the other hand, those poses where $T_{\textrm{map} \leftarrow \textrm{agent}}$ is known and accurately optimized in bundle-adjustment are already old and the odometry may have drifted. We therefore run a sliding-window average over the most recent poses where we know both $T_{\textrm{map} \leftarrow \textrm{agent}}$ and $T_{\textrm{odom} \leftarrow \textrm{agent}}$ and use this average to integrate and transform the odometry into a global robot/HoloLens pose. In particular, we note that this is more stable than estimating $T_{\textrm{map} \leftarrow \textrm{odom}}$ from the most recent node in the global pose graph, because we can adjust the trade off between bundle-adjustment and odometry drift through the size of the sliding window.
\begin{align*}
    T_{\textrm{map} \leftarrow \textrm{agent}}^{(t-0)} = \Avg\limits_{t-i \leq t < t-j} \left( T_{\textrm{map} \leftarrow \textrm{agent}}^{(t)} T_{\textrm{agent} \leftarrow \textrm{odom}}^{(t)} \right) T_{\textrm{odom} \leftarrow \textrm{agent}}^{(t-0)}
\end{align*}
We illustrate the approach in Figure~\ref{fig:frame-management}.

\subsubsection{Dense Reconstruction}
For all those agents that have an RGB-D sensor, we regularly integrate all RGB-D frames and their optimized global poses into a common 3D dense reconstruction using voxblox~\cite{8202315}. This yields textured surface meshes and respectively colored surface point clouds that we use for visualization. It also yields an SDF representation of the environment for planning.

\subsubsection{Map Overlay Visualization in HoloLens}
After the HoloLens and robot agents are co-localized, we can display the 3D surface point cloud map in the HoloLens, overlaid as a hologram onto the real world. An example of this perspective can be found in Fig. \ref{fig:result}. Because the HoloLens is localized in the global map, we are able to display the 3D point cloud in the correct orientation in the Unity app by performing the proper coordinate transforms---from the global map frame in the server, to the HoloLens visualization frame. 

When implementing this functionality, we noticed that delays between the map optimization and the self-odometry estimation of the device can cause rotational misalignment. These are not severe, but in large environments even 1 degree error in the orientation can cause severe displacements of the visualized map a few meters away from the visualization origin. Therefore, when transforming the global map into the visualization origin of the HoloLens, we take position and yaw from the global reference frame management, but update roll and pitch with the newest estimation of the gravity direction on the HoloLens. This leads to much more stable visualizations. Similar to the global map, we transform the current pose of all robots into the HoloLens visualization frame and display a hologram marker tracking their pose.


\subsection{Drag-and-Drop Robot Control Interface}
\label{subsec:mr-robot-collab}

To control the robots by dragging and dropping their holograms, like shown in Figure~\ref{fig:arche_controller_mode}, we switch the map visualization into a `mini map' mode that is scaled down to bring all parts of the map within hand's reach of the operator.

\subsubsection{Drag-and-Drop Interface} In this interface, we represent each robot with 2 holograms: One tracking the current position of the robot, marked with a round halo above and a rectangular plate under the robot, and a second hologram that can be manipulated by the user. Whenever the second hologram is released, its 6DoF pose in the map is sent to the reference frame manager, and translated into a relative local goal pose in the respective robot's odometry frame.

\subsubsection{Viability Check and Path Planning} Depending on the robot model, different poses are viable and reachable: An omnidirectional drone may reach any pose in free space outside of its collision radius. A ground robot may only reach poses on a flat surface. 
A quadrupedal robot can reach poses at a certain height above ground, but also traverse stairs and more rough terrain. Instead of requiring a user to select only goal poses that satisfy all constraints, we add a central ``viability" and ``correction" step before sending out goal poses to the robots. Thus in the global map, we estimate ground surfaces and collision volumes from the global SDF map and compare them against the user's sent goal pose. Then, there are 3 options:
\begin{itemize}
    \item If the goal pose is completely viable, we simply forward it to the robot
    \item If the goal pose is not viable, but we find a viable pose within a certain radius, we forward that viable pose
    \item If the goal pose is not viable and we cannot find a nearby viable pose, nothing is forwarded
\end{itemize}
The correction step makes it easier to e.g. command a robot into a corridor that is only marginally wider than its collision volume. This decreases the need for manual correction and gives operators more leeway when giving commands.

Once a viable goal pose is found, a light-weight planning module gets activated. It runs a 3D RRT \cite{LaValle1998RapidlyexploringRT} path planner utilizing the SDF obtained from the global map. It builds a trajectory of waypoints between the current position and the goal, and checks that all waypoints and connections between them are traversable. We focus our experiments on walking robots, but the same method applies to drones or wheeled robots as they have simpler traversability constraints. To estimate quadruped traversability of waypoints, we check that poses are within a height corridor above a ground plane and set a maximum height difference between any two connected waypoints. Subsequently, the planned path is sent to the robot, transformed into its local frame, and followed by a local, obstacle avoiding planning and control system.

\section{User Study}
To evaluate the effectiveness of our proposed drag-and-drop interface, we follow the approach of~\cite{10161412} and conduct a comparative user study between our newly proposed interface and the established RViz interface from ROS. Our hypothesis is that it is easier for users to understand a 3D multi-floor space and set goal poses in an MR device. Therefore, the study tests the ease of examining a 3D point cloud map and then using the interface to send a goal pose to a robot. As a consequence, every study participant sees only one of the two interfaces, since the environment is already known and understood after completing the tasks once. The study was approved by the ETH Ethics Commission.

\subsection{Study Setup}
The first four tasks involve operating a fleet of three robots in a 3D map of a two-story home. The fifth task requires users to explore an unmapped space. The 5 tasks are always assigned in the same order. Task 1 is to ``send one of the robots to the large upstairs bedroom." Task 2 is to ``send one robot to each bedroom and a third robot to the living room." Task 3 is to ``switch the poses of the red and blue robot." Task 4 is to ``gather all robots around the kitchen table." For tasks 1 and 2, all robots start at the building entrance. For tasks 3 and 4, they are always initialized at the same various locations within the building. Finally, for Task 5, we change the environment to a one-story apartment, and users just see one initial room. Task 5 is to ``find a bedroom by exploring and mapping the apartment with the robot."

For each task, we measure the time users take to set all goal positions, as well as the time until all robots have reached their goals. The second time is dependent on the planning and execution, but also the coordination of the user to operate multiple robots. For example, moving the furthest robot first allows it more time to get to its goal, since robots can execute tasks asynchronously. Additionally, we ask users to fill out the NASA TLX form~\cite{nasatlx} after every task. This is a standard test to measure cognitive and physical load and is also used in related work~\cite{10161412}. To ensure that all robot positions, plans, and controls are repeatable and do not interfere with the study, the interfaces are connected to NVIDIA Isaac Simulator, instead of physical robots.

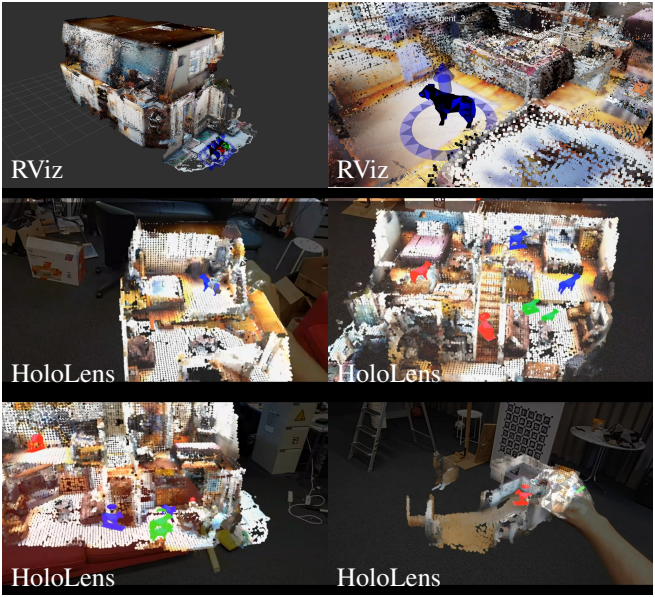
\begin{figure}
\centering
\begin{tikzpicture}
    \node[inner sep=0] (1) {\includegraphics[width=0.5\linewidth]{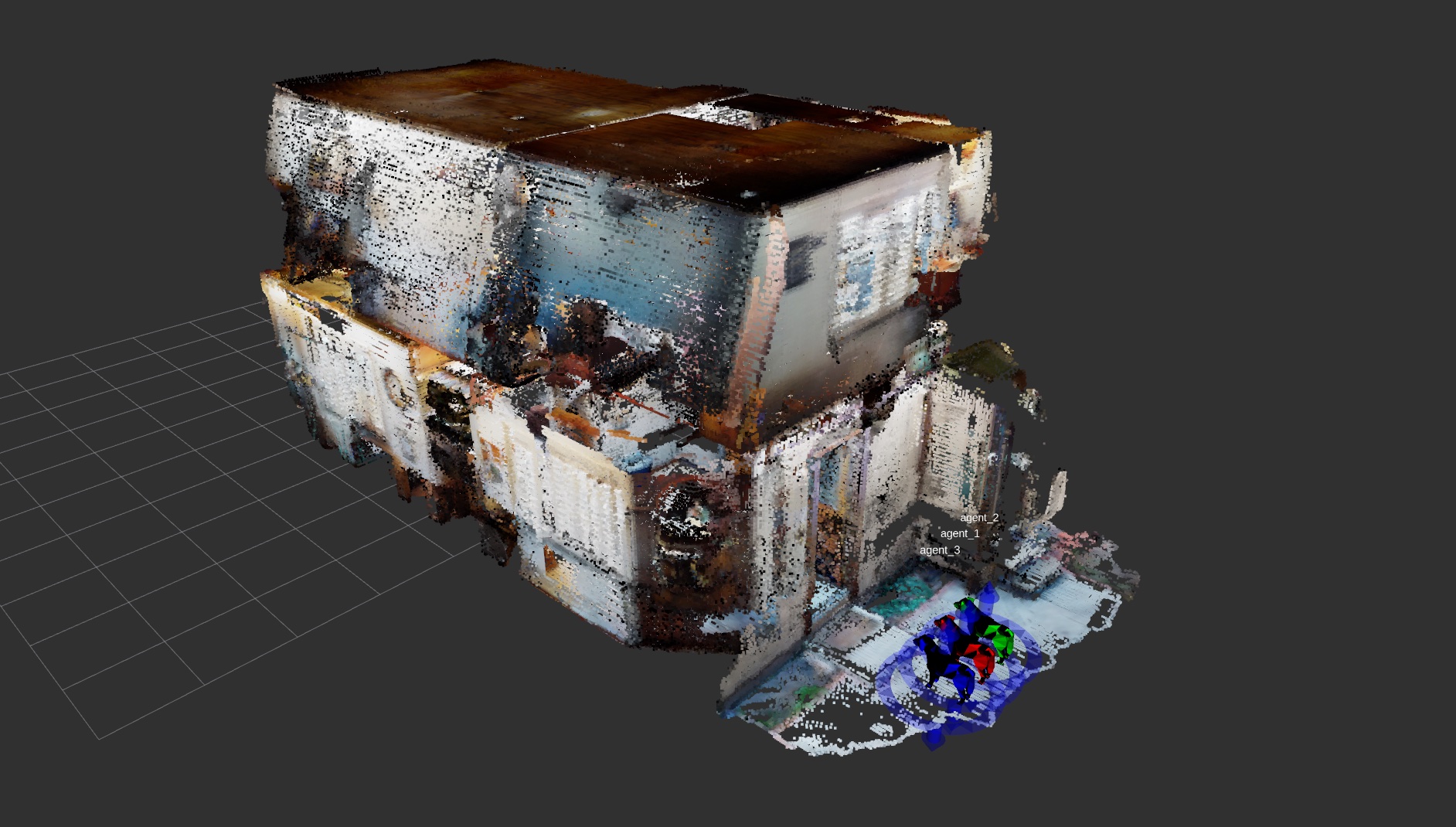}};
    \node[right=0pt of 1,inner sep=0] (2) {\includegraphics[width=0.5\linewidth]{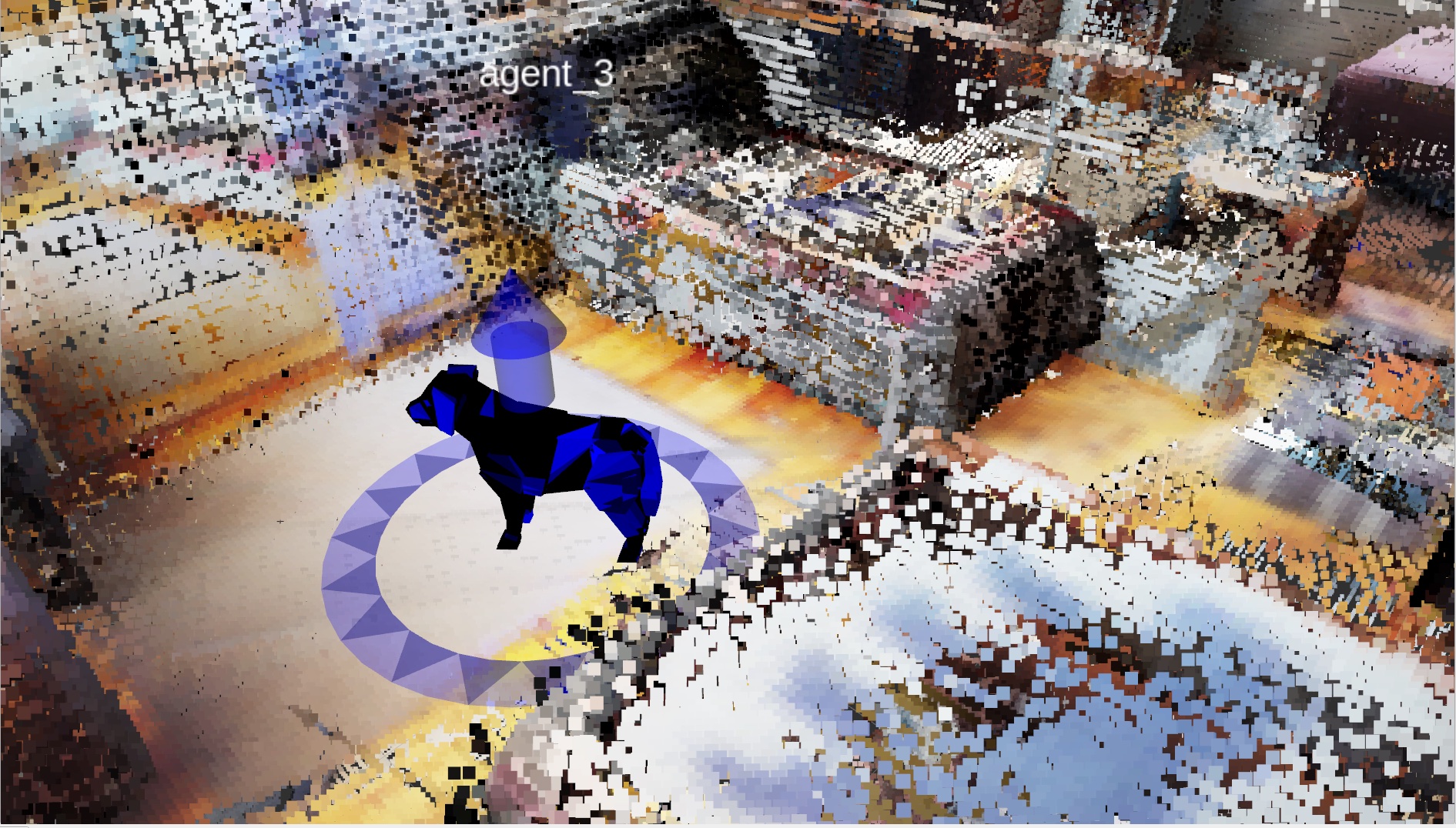}};
    \node[below=0pt of 1,inner sep=0] (3) {\includegraphics[width=0.5\linewidth]{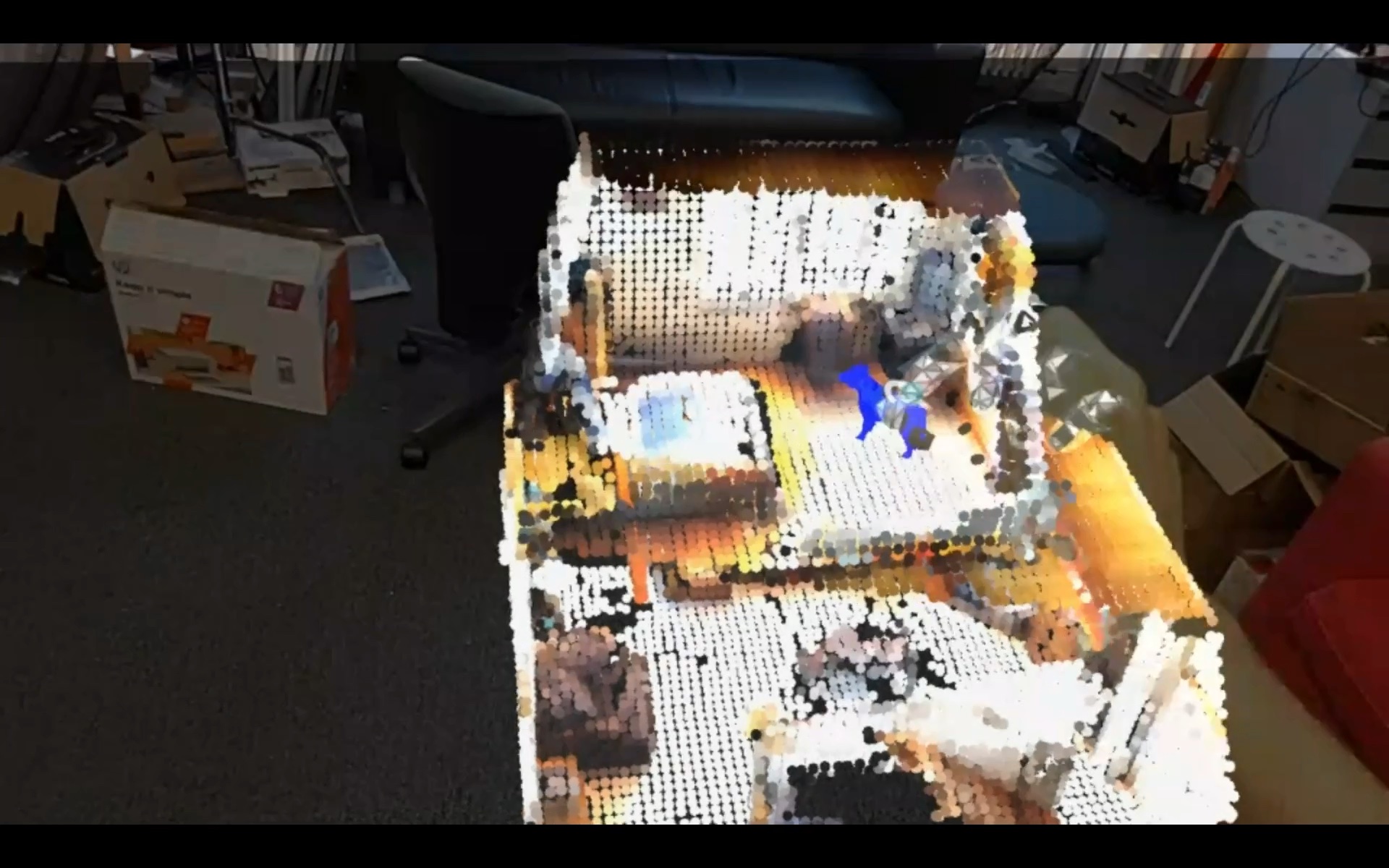}};
    \node[right=0pt of 3,inner sep=0] (4) {\includegraphics[width=0.5\linewidth]{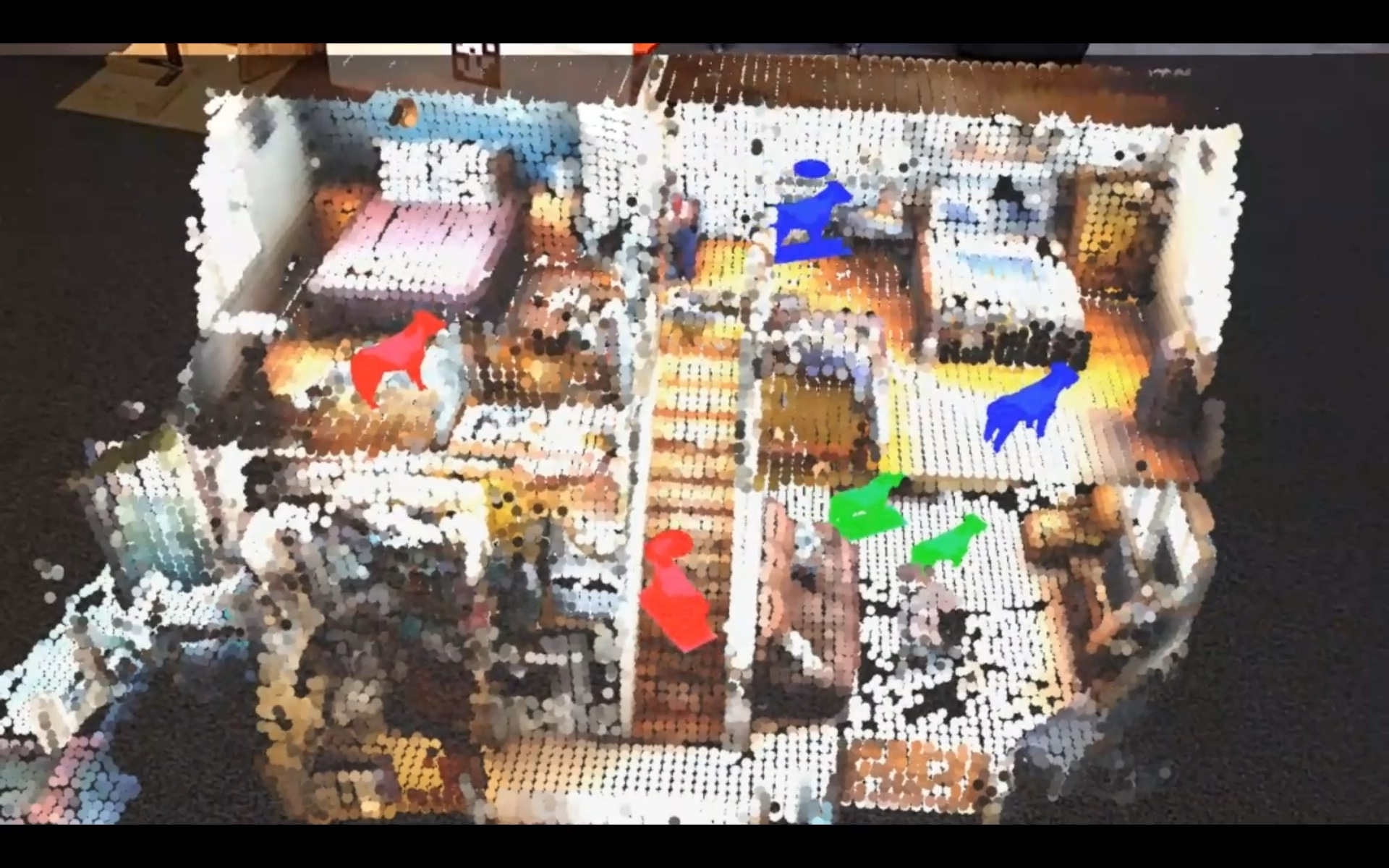}};
    \node[below=0pt of 3,inner sep=0] (5) {\includegraphics[width=0.5\linewidth]{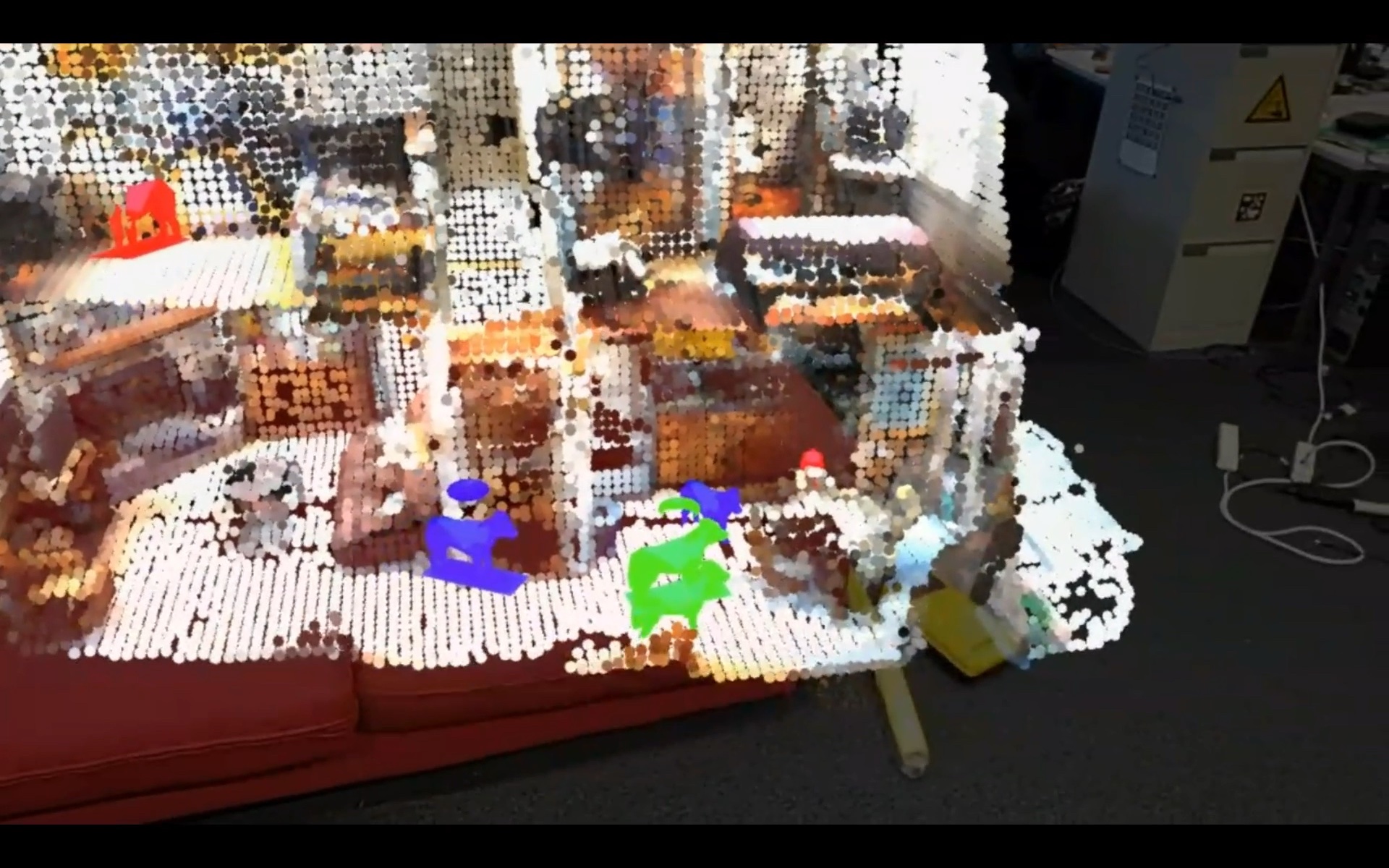}};
    \node[right=0pt of 5,inner sep=0] (6) {\includegraphics[width=0.5\linewidth]{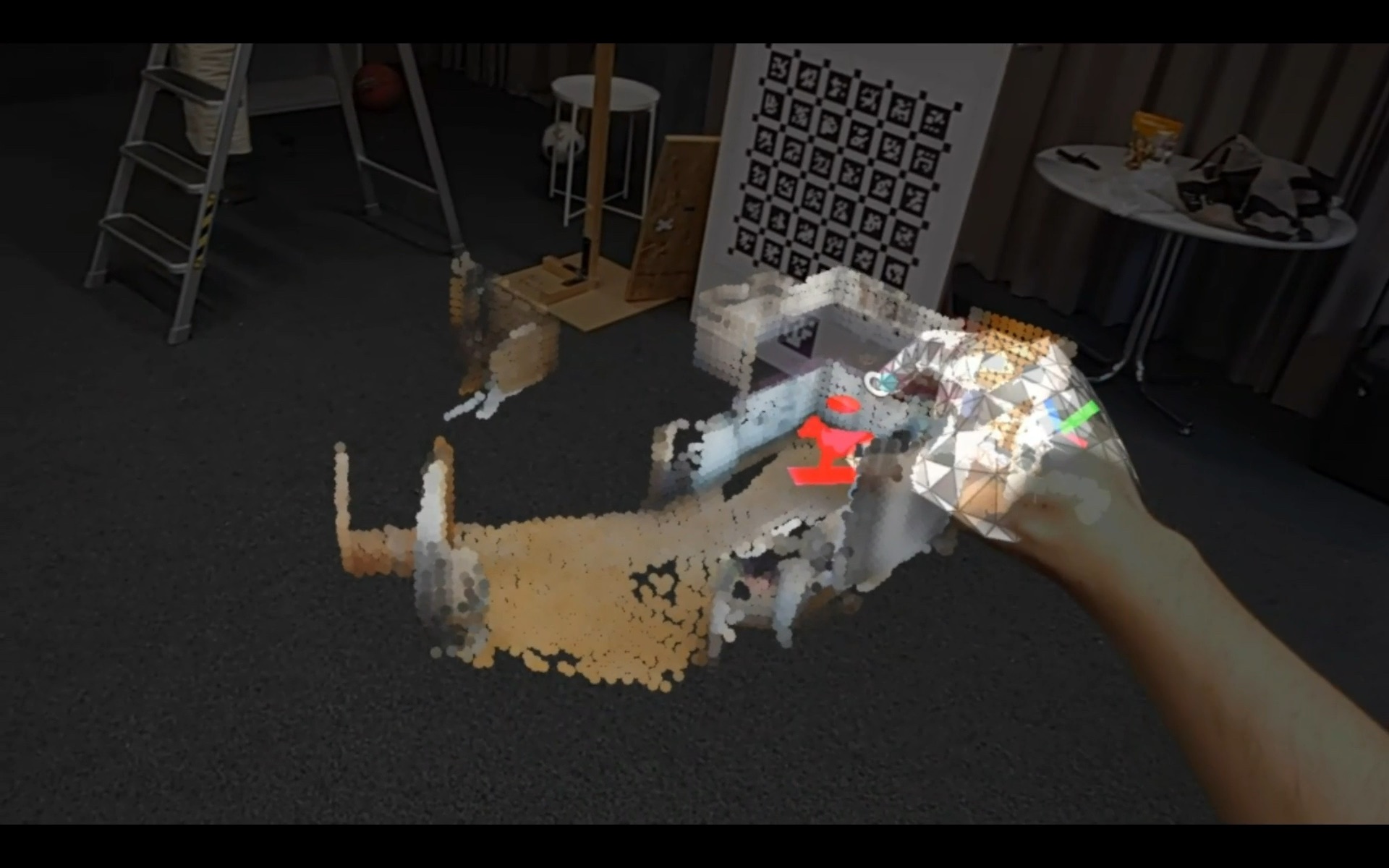}};
    \node[anchor=south west] at (1.south west) {\color{white} RViz};
    \node[anchor=south west] at (2.south west) {\color{white} RViz};
    \node[anchor=south west] at (3.south west) {\color{white} HoloLens};
    \node[anchor=south west] at (4.south west) {\color{white} HoloLens};
    \node[anchor=south west] at (5.south west) {\color{white} HoloLens};
    \node[anchor=south west] at (6.south west) {\color{white} HoloLens};
\end{tikzpicture}
    \caption{\textbf{User Study Setup} In the top left we show the two-story environment for tasks 1-4. In the top right and middle left, we show an agent being placed in the larger bedroom (Task 1). The middle right (Task 2) and bottom left (Task 4) show three agents navigating to their respective goal poses. And the bottom right shows the environment for Task 5 where users continue to map the space as they explore. A better view of the environment for tasks 1-4 can be found \href{https://aihabitat.org/datasets/hm3d/00164-XfUxBGTFQQb/index.html}{here}, and for Task 5 \href{https://aihabitat.org/datasets/hm3d/00108-oStKKWkQ1id/index.html}{here}.}
    \label{fig:task_ucug}
    \vspace{-6mm}
\end{figure}

As a baseline interface, we use an RViz visualization of the same 3D maps as shown in our HoloLens interface. Robot goal poses are controlled in RViz by dragging interactive markers around on a plane, or by shifting the height along a vertical line, as shown in the top right images of Fig.~\ref{fig:task_ucug}. 

Before starting with Task 1, we let users familiarize themselves with the controls in an empty plane environment. Users practice how to control their viewing angle, i.e. changing viewpoints in RViz or physically moving around the hologram with the HoloLens. Users also practice commanding robots to different poses. We only proceed to Task 1 once users confirm they are ready.

\begin{figure*}
\centering
\includegraphics[width=\linewidth]{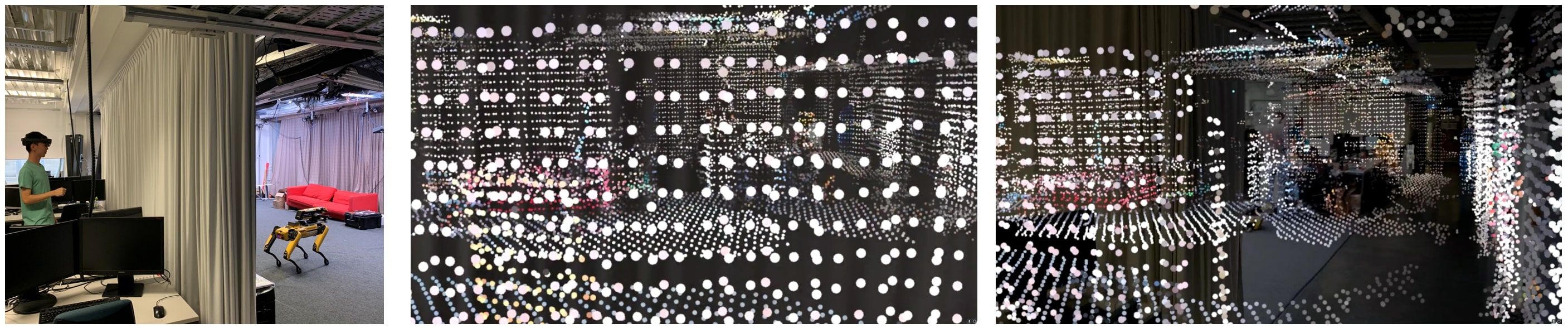}
\caption{\textbf{Result of Map Overlay} The left image shows the corresponding environment for the middle and right images. After co-localizing the HoloLens and the robot, the operator is able to ``see through" the curtain into the space behind, because the robot mapped it. The middle and right images are first-person perspectives of what the operator sees through the HoloLens.}
\label{fig:result}
\vspace{-5mm}
\end{figure*}

\subsection{Study Results}

\begin{figure}
    \includegraphics[width=\linewidth]{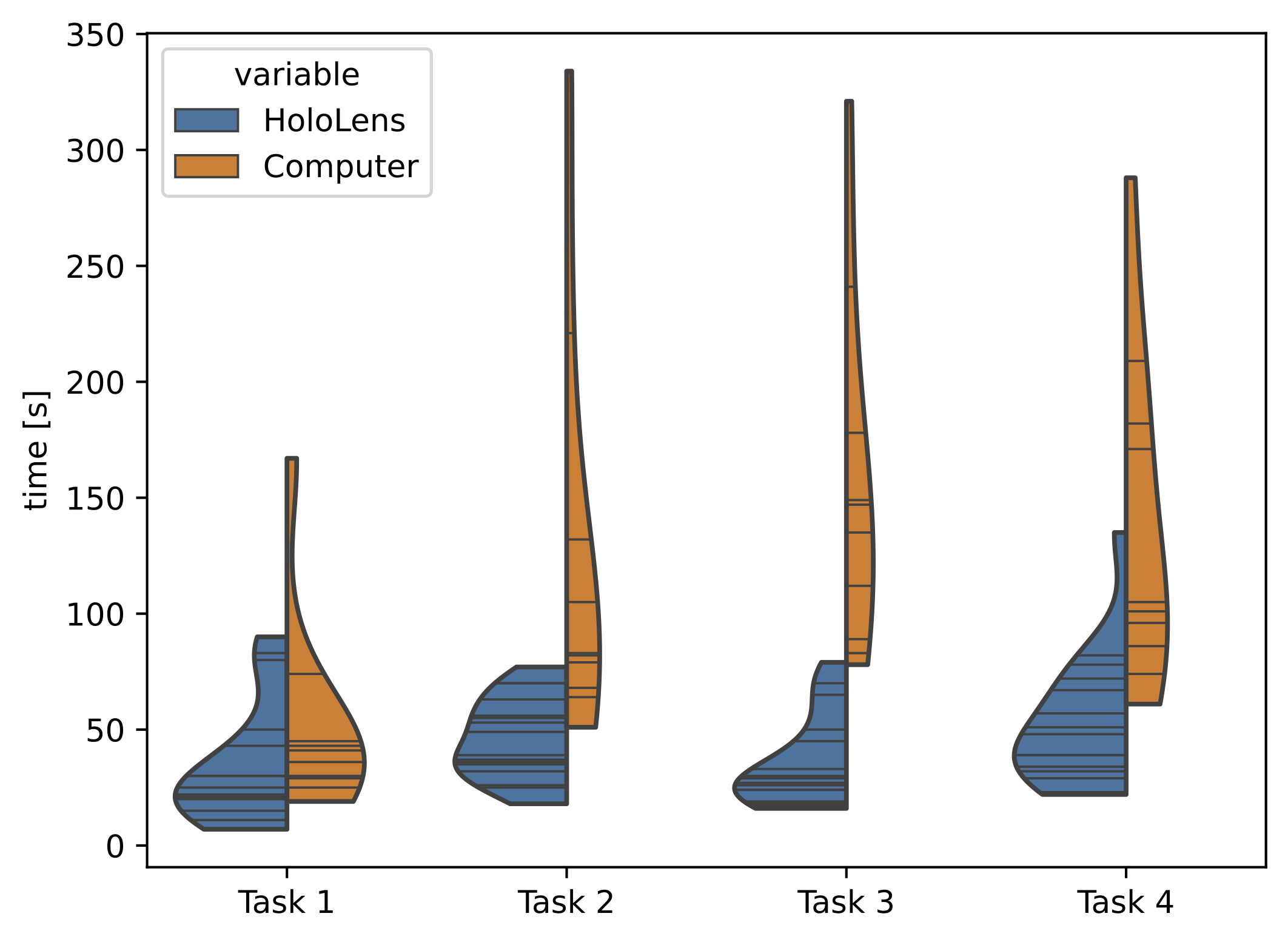}
    \caption{\textbf{Time Until Goal} for tasks 1-4. This is the time it took users to finish giving commands to the robots. This time does not include the remaining time it took for the robot to actually navigate to their final positions. Based on a one-tailed Mann Whitney U Test~\cite{wilcoxon} there is a significant ($\alpha =0.01$) difference between devices for tasks 2-4.
    }  
    \label{fig:task_ug}
    \vspace{-5mm}
\end{figure}

\begin{figure}
    \includegraphics[width=\linewidth]{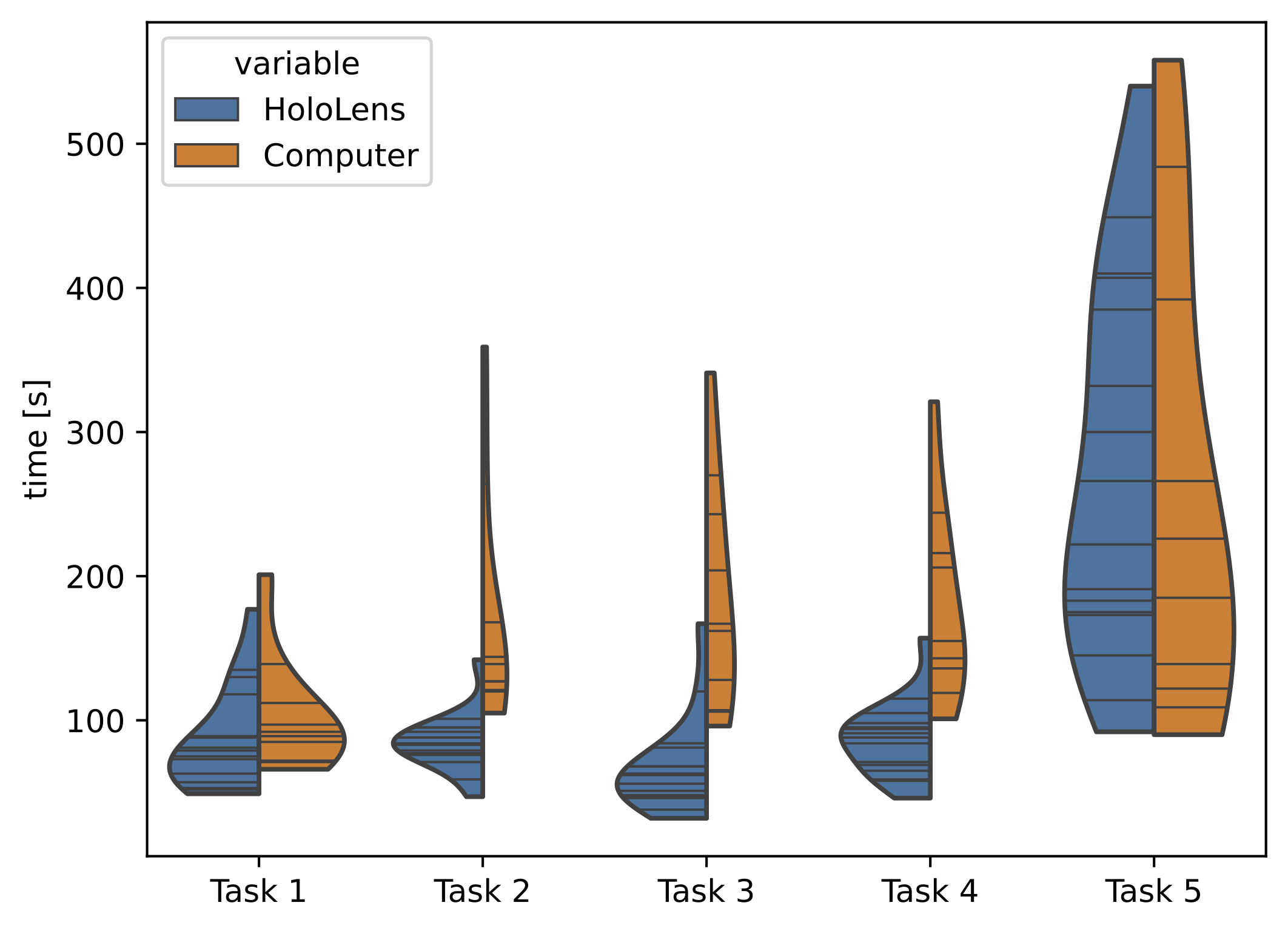}
    \caption{\textbf{Time Until Completion} for tasks 1-5. This is the time until all agents have moved to their final positions (tasks 1-4), or until the user has found a bedroom (Task 5). We perform a one-tailed Mann Whitney U Test~\cite{wilcoxon} which shows a statistical significance ($\alpha =0.01$) for the performance time distributions for tasks 2-4. Users have similar performance times for tasks 1 and 5 with both interfaces.} 
    \label{fig:task_uc}
    \vspace{-5mm}
\end{figure}

Our study results are illustrated in Fig.~\ref{fig:task_ug} and Fig.~\ref{fig:task_uc}.
Following the approach of~\cite{10161412}, we first check if the data is normally distributed by conducting a Shapiro-Wilk test~\cite{shapiro1965analysis}, which indicates that we need to use a non-parametric test to compare the user performance times between devices. To determine if our data between devices comes from the same or different distributions, we use the one-tailed Mann Whitney U Test~\cite{wilcoxon}. 
For tasks 2-4, we measure a significant ($\alpha = .01$) difference between the medians of the HoloLens and RViz performance times. For tasks 1 and 5, we measure no statistical significance in the difference between the median times. We also perform a statistical test on the TLX scores. For $\alpha = .05$, only Task 3 has a statistical significance for lower HoloLens TLX scores. We had 16 users test the HoloLens interface, and 10 other users for RViz.

From these results, we posit that the HoloLens is a better interface for performing more complex robot control tasks with multiple robots.
Tasks 2-4 differ from 1 and 5 as they include multiple robots and ask operators to look around in an enclosed point cloud of a two-story building. Tasks 1 and 5 are easier in the sense that they both have only one agent.
The design of Task 5 was for the purpose of determining how well users are able to understand a 3D environment through the different interfaces. At least for a single-story single-robot environment, our HoloLens interface was not better or worse than RViz. Given the results for tasks 2-4, future work will need to investigate whether this changes if the environment becomes more complex, and/or if users can control multiple robots for one exploration task.

There are no statistically significant differences for the TLX scores between devices, except for Task 3. We posit that this could be due to the fact that this task requires a precise ending goal position and orientation. The HoloLens interface offers the advantage that changing the position and orientation is as simple as drag and drop, whereas RViz requires multiple steps to first change the height of the robot, then position and orientation, or vice versa. 

During the experiments we also collected qualitative impressions. Some users mentioned that changing their viewpoint by moving around and into the hologram is more intuitive than what they knew from other interfaces. However, we noticed that many HoloLens users had difficulties with the drag-and-drop action due hand tracking failures, similar to~\cite{10161412}. For RViz, we noticed some users having difficulties with zooming and panning the viewpoint to an appropriate perspective for performing the task, which is a natural limitation of navigating a 3D scene on a 2D screen.

\section{Real-World Tests}
In addition to the user study that validates the effectiveness of our drag-and-drop interface, but uses simulated robots to ensure repeatability of the tasks, we validate the functionality of the full proposed system in real-world experiments.

For this, we connect the HoloLens 2 to a Boston Dynamics Spot quadruped with an RGB-D camera and onboard compute for local planning and mapping. An external workstation runs the global mapping and reference frame management. Because the HoloLens has limited compute and does not provide access to its hardware-accelerated internal mapping, we stream raw sensor data over WiFi and run the local mapping of the HoloLens also on an external workstation.
\label{experiment-setup}

\subsection{Map Overlay Visualization}
We show the map overlay, achieved through co-localization and reference frame transformation, in Figure~\ref{fig:result}. Note that the HoloLens does not offer any interface to measure where it visualizes 3D holograms with respect to its sensor rig or outside physical references, making it difficult to quantify if a hologram is placed correctly. Our validation is therefore reduced to these qualitative results. As demonstrated in Fig. \ref{fig:result} and the supplemental video, the overlay visualization in the HoloLens displays the 3D point cloud from the map \textit{accurately overlaid} onto the real-world environment, within a maximum offset of ${\sim}20cm$. We therefore achieve a similar interface as~\cite{erat2018drone,reardon2018come}, but fully automatically through global visual registration, and with a live updated 3D map representation, whereas \cite{erat2018drone} requires a human to pre-model a mesh of the space and \cite{reardon2018come} shows a 2D map.
This, as visualized, allows the operator to ``see through'' the curtain, enabling them to easily supervise the robot deployment even into previously unknown environments.


\subsection{Drag-and-Drop Interface}
Figure~\ref{fig:arche_controller_mode} and the accompanying video show the mini-map 3D robot control interface, which allows operators to directly give commands to the robot in 3D space. We demonstrate how the operator is then able to move the robot through a doorway, as well as up and down stairs. In 4 out of 5 trials we conducted in the field, the operator successfully controlled the robot through the doorway, up the stairs, and out of the building again, while in 1 trial an issue with the local controller (outside of our system) caused the robot to fall on the step into the building. While the user study validated the advantages of the drag-and-drop interface, this experiment validates that the approach can be deployed to a real environment and used to successfully operate a quadrupedal robot.

\section{Conclusion and Future Work}
We introduce a novel mixed-reality human-robot teaming system. This system offers multi-agent mapping and localization capabilities while providing operators, wearing MR devices, with the ability to visualize outcomes and send commands to robots via the MR interface. A user study shows that users are significantly faster at commanding a team of robots in 3D, compared to the standard RViz interface. Through real-world experiments, we showcase the system's applicability and potential for diverse applications, including but not limited to search and rescue operations, industrial inspections, and more. Our work underscores the versatility and effectiveness of MR interfaces in enhancing human-robot teaming in practical scenarios.
There are many exciting directions for future work, such as extending the current system towards a multi-robot autonomous and human-guided exploration settings.




\bibliographystyle{ieee}
\bibliography{mybibliography}

\begin{thebibliography}{10}\itemsep=-1pt

\bibitem{teleop3_bejczy2020_mr_interface_for_improving_mob_manip}
B.~Bejczy, R.~Bozyil, E.~Vaičekauskas, S.~B.~K. Petersen, S.~Bøgh, S.~S.
  Hjorth, and E.~B. Hansen.
\newblock Mixed reality interface for improving mobile manipulator
  teleoperation in contamination critical applications.
\newblock {\em Procedia Manufacturing}, 51:620--626, 2020.

\bibitem{Cramariuc-2023}
A.~Cramariuc, L.~Bernreiter, F.~Tschopp, M.~Fehr, V.~Reijgwart, J.~Nieto,
  R.~Siegwart, and C.~Cadena.
\newblock maplab 2.0 {\textendash} a modular and multi-modal mapping framework.
\newblock {\em {IEEE} Robotics and Automation Letters}, 8(2):520--527, feb
  2023.

\bibitem{cruz2023mixed}
C.~Cruz~Ulloa, J.~del Cerro, and A.~Barrientos.
\newblock Mixed-reality for quadruped-robotic guidance in sar tasks.
\newblock {\em Journal of Computational Design and Engineering},
  10(4):1479--1489, 2023.

\bibitem{9681715}
J.~Delmerico, R.~Poranne, F.~Bogo, H.~Oleynikova, E.~Vollenweider, S.~Coros,
  J.~Nieto, and M.~Pollefeys.
\newblock Spatial computing and intuitive interaction: Bringing mixed reality
  and robotics together.
\newblock {\em IEEE Robotics \& Automation Magazine}, 29(1):45--57, 2022.

\bibitem{erat2018drone}
O.~Erat, W.~A. Isop, D.~Kalkofen, and D.~Schmalstieg.
\newblock Drone-augmented human vision: Exocentric control for drones exploring
  hidden areas.
\newblock {\em IEEE transactions on visualization and computer graphics},
  24(4):1437--1446, 2018.

\bibitem{sutd_video}
F.R.I.
\newblock Online robot path planning via virtual map on hololens2, 2022.
\newblock YouTube video.

\bibitem{nasatlx}
S.~G. Hart and L.~E. Staveland.
\newblock Development of nasa-tlx (task load index): Results of empirical and
  theoretical research.
\newblock In {\em Advances in psychology}, volume~52, pages 139--183. Elsevier,
  1988.

\bibitem{wilcoxon}
W.~Haynes.
\newblock {\em Wilcoxon Rank Sum Test}, pages 2354--2355.
\newblock 01 2013.

\bibitem{hedayati2018improving}
H.~Hedayati, M.~Walker, and D.~Szafir.
\newblock Improving collocated robot teleoperation with augmented reality.
\newblock In {\em Proceedings of the 2018 ACM/IEEE International Conference on
  Human-Robot Interaction}, pages 78--86, 2018.

\bibitem{9561105}
F.~Kennel-Maushart, R.~Poranne, and S.~Coros.
\newblock Manipulability optimization for multi-arm teleoperation.
\newblock In {\em 2021 IEEE International Conference on Robotics and Automation
  (ICRA)}, pages 3956--3962, 2021.

\bibitem{10161412}
F.~Kennel-Maushart, R.~Poranne, and S.~Coros.
\newblock Interacting with multi-robot systems via mixed reality.
\newblock In {\em 2023 IEEE International Conference on Robotics and Automation
  (ICRA)}, pages 11633--11639, 2023.

\bibitem{LaValle1998RapidlyexploringRT}
S.~M. LaValle.
\newblock Rapidly-exploring random trees : a new tool for path planning.
\newblock {\em The annual research report}, 1998.

\bibitem{teleop5_rob_teleop_based_on_mr}
C.~Liang, C.~Liu, X.~Liu, L.~Cheng, and C.~Yang.
\newblock Robot teleoperation system based on mixed reality.
\newblock In {\em 2019 IEEE 4th International Conference on Advanced Robotics
  and Mechatronics (ICARM)}, pages 384--389, 2019.

\bibitem{mott2021you}
T.~Mott, T.~Williams, H.~Zhang, and C.~Reardon.
\newblock You have time to explore over here!: Augmented reality for enhanced
  situation awareness in human-robot collaborative exploration.
\newblock In {\em 4th International Workshop on Virtual, Augmented, and Mixed
  Reality for HRI}, 2021.

\bibitem{naceri2021vicarios}
A.~Naceri, D.~Mazzanti, J.~Bimbo, Y.~T. Tefera, D.~Prattichizzo, D.~G.
  Caldwell, L.~S. Mattos, and N.~Deshpande.
\newblock The vicarios virtual reality interface for remote robotic
  teleoperation: teleporting for intuitive tele-manipulation.
\newblock {\em Journal of Intelligent \& Robotic Systems}, 101:1--16, 2021.

\bibitem{8202315}
H.~Oleynikova, Z.~Taylor, M.~Fehr, R.~Siegwart, and J.~Nieto.
\newblock Voxblox: Incremental 3d euclidean signed distance fields for on-board
  mav planning.
\newblock In {\em 2017 IEEE/RSJ International Conference on Intelligent Robots
  and Systems (IROS)}, pages 1366--1373, 2017.

\bibitem{reardon2020enabling}
C.~Reardon, J.~Gregory, C.~Nieto-Granda, and J.~G. Rogers.
\newblock Enabling situational awareness via augmented reality of autonomous
  robot-based environmental change detection.
\newblock In {\em Virtual, Augmented and Mixed Reality. Design and Interaction:
  12th International Conference, VAMR 2020, Held as Part of the 22nd HCI
  International Conference, HCII 2020, Copenhagen, Denmark, July 19--24, 2020,
  Proceedings, Part I 22}, pages 611--628. Springer, 2020.

\bibitem{reardon2018come}
C.~Reardon, K.~Lee, and J.~Fink.
\newblock Come see this! augmented reality to enable human-robot cooperative
  search.
\newblock In {\em 2018 IEEE International Symposium on Safety, Security, and
  Rescue Robotics (SSRR)}, pages 1--7. IEEE, 2018.

\bibitem{reardon2019communicating}
C.~Reardon, K.~Lee, J.~G. Rogers, and J.~Fink.
\newblock Communicating via augmented reality for human-robot teaming in field
  environments.
\newblock In {\em 2019 IEEE International Symposium on Safety, Security, and
  Rescue Robotics (SSRR)}, pages 94--101. IEEE, 2019.

\bibitem{teleop2_sauer2009_towards_a_predictive_mr_ui_for_mobile_robot_teleop}
M.~Sauer, M.~Hess, and K.~Schilling.
\newblock Towards a predictive mixed reality user interface for mobile robot
  teleoperation.
\newblock {\em IFAC Proceedings Volumes}, 42(22):91--96, 2009.

\bibitem{teleop4_sauer2010_mr_ui_for_mobile_robot_teleop_in_adhoc}
M.~Sauer, F.~Zeiger, and K.~Schilling.
\newblock Mixed-reality user interface for mobile robot teleoperation in ad-hoc
  networks.
\newblock {\em IFAC Proceedings Volumes}, 43(23):77--82, 2010.

\bibitem{shapiro1965analysis}
S.~S. Shapiro and M.~B. Wilk.
\newblock An analysis of variance test for normality (complete samples).
\newblock {\em Biometrika}, 52(3/4):591--611, 1965.

\bibitem{8968598}
P.~Stotko, S.~Krumpen, M.~Schwarz, C.~Lenz, S.~Behnke, R.~Klein, and
  M.~Weinmann.
\newblock A vr system for immersive teleoperation and live exploration with a
  mobile robot.
\newblock In {\em 2019 IEEE/RSJ International Conference on Intelligent Robots
  and Systems (IROS)}, pages 3630--3637, 2019.

\bibitem{Suzuki_2022}
R.~Suzuki, A.~Karim, T.~Xia, H.~Hedayati, and N.~Marquardt.
\newblock Augmented reality and robotics: A survey and taxonomy for
  {AR}-enhanced human-robot interaction and robotic interfaces.
\newblock In {\em {CHI} Conference on Human Factors in Computing Systems}.
  {ACM}, apr 2022.

\bibitem{teleop6_legged_manip}
C.~Ulloa~C, D.~Domínguez, J.~Del~Cerro, and A.~Barrientos.
\newblock A mixed-reality tele-operation method for high-level control of a
  legged-manipulator robot.
\newblock {\em Sensors (Basel)}, 22(21):8146, 2022.

\bibitem{walker2022virtual}
M.~Walker, T.~Phung, T.~Chakraborti, T.~Williams, and D.~Szafir.
\newblock Virtual, augmented, and mixed reality for human-robot interaction: A
  survey and virtual design element taxonomy.
\newblock {\em ACM Transactions on Human-Robot Interaction}, 2022.

\bibitem{teleop1_robot_teleop_ar_virtual_surrogates}
M.~E. Walker, H.~Hedayati, and D.~Szafir.
\newblock Robot teleoperation with augmented reality virtual surrogates.
\newblock In {\em 2019 14th ACM/IEEE International Conference on Human-Robot
  Interaction (HRI)}, pages 202--210, 2019.

\bibitem{wu2018omnidirectional}
M.~Wu, Y.~Xu, C.~Yang, and Y.~Feng.
\newblock Omnidirectional mobile robot control based on mixed reality and semg
  signals.
\newblock In {\em 2018 Chinese Automation Congress (CAC)}, pages 1867--1872.
  IEEE, 2018.

\end{thebibliography}

\end{document}